\documentclass[english]{article}
\usepackage[T1]{fontenc}
\usepackage{textcomp}
\usepackage[latin9]{inputenc}
\usepackage{amstext}

\makeatletter

\providecommand{\tabularnewline}{\\}

\makeatother

\usepackage{babel}
\begin{document}
\title{On some improvements to Unbounded Minimax}
\author{Quentin Cohen-Solal and Tristan Cazenave\\
LAMSADE, Université Paris-Dauphine, PSL, CNRS\\
Paris, France\\
quentin.cohen-solal@dauphine.psl.eu and tristan.cazenave@lamsade.dauphine.fr}
\maketitle
\begin{abstract}
This paper presents the first experimental evaluation of four previously
untested modifications of Unbounded Best-First Minimax algorithm.
This algorithm explores the game tree by iteratively expanding the
most promising sequences of actions based on the current partial game
tree. We first evaluate the use of transposition tables, which convert
the game tree into a directed acyclic graph by merging duplicate states.
Second, we compare the original algorithm by Korf \& Chickering with
the variant proposed by Cohen-Solal, which differs in its backpropagation
strategy: instead of stopping when a stable value is encountered,
it updates values up to the root. This change slightly improves performance
when value ties or transposition tables are involved. Third, we assess
replacing the exact terminal evaluation function with the learned
heuristic function. While beneficial when exact evaluations are costly,
this modification reduces performance in inexpensive settings. Finally,
we examine the impact of the completion technique that prioritizes
resolved winning states and avoids resolved losing states. This technique
also improves performance. Overall, our findings highlight how targeted
modifications can enhance the efficiency of Unbounded Best-First Minimax.
\end{abstract}
\noindent\begin{minipage}[t]{1\columnwidth}%
\global\long\def\ubfmref{\text{\ensuremath{\mathrm{UBFM_{ref}}}}}%

\global\long\def\ubfmnoc{\ensuremath{\mathrm{\ubfmref^{\setminus C}}}}%

\global\long\def\ubfmnocete{\ensuremath{\mathrm{\ubfmref^{\setminus C\setminus f_{\mathrm{t}}}}}}%

\global\long\def\ubfmnott{\ensuremath{\mathrm{\ubfmref^{\setminus TT}}}}%

\global\long\def\ubfmkcback{\ensuremath{\mathrm{\ubfmref^{B_{K\&C}}}}}%

\global\long\def\ubfmnoete{\ensuremath{\ubfmref^{\setminus f_{\mathrm{t}}}}}%
\end{minipage}

\section{Introduction}

Recent years have seen remarkable progress in game-playing algorithms,
most notably with the development of AlphaZero, which achieved superhuman
performance in games such as chess and Go through reinforcement learning,
without relying on any human domain knowledge.~\cite{silver2018general,silver2017mastering}.
AlphaZero\textquoteright s impact has been profound, both in terms
of performance and its influence on subsequent research directions.
In particular, the implications of this algorithm go well beyond games:
its application has notably made it possible to identify a more efficient
sorting algorithm \cite{mankowitz2023faster}. as well as a more efficient
matrix multiplication algorithm~ \cite{fawzi2022discovering}.

More recently, a new algorithm named Athénan has been introduced~\cite{cohen2020learning},
which has demonstrated significant advantages over AlphaZero. Experimental
results have shown that Athénan is not only more efficient---being
over 30 times faster---but also more performant in certain games,
even when AlphaZero is allocated 100 times more computational resources
(e.g., 100 GPUs)~\cite{cohen2023minimax}. 

At the core of Athénan lies the Unbounded Best-First Minimax (UBFM)
algorithm~\cite{cohen2025study,cohen2020learning,korf1996best},
a search algorithm that had received little attention prior to Athénan\textquoteright s
introduction. Unlike traditional Alpha-Beta pruning, UBFM performs
a best-first expansion of action sequences, progressively deepening
its understanding of the game state space. A recent empirical study
has demonstrated that UBFM, when combined with the safe decision technique,
is the best-performing search algorithm under short time constraints
and also achieves the best average performance in medium time conditions~\cite{cohen2025study}.
These results position UBFM as a promising alternative to traditional
search strategies in adversarial games.

In this work, we conduct the first experimental study of four modifications
to the original UBFM algorithm. Although these modifications have
been proposed in a prior article~\cite{cohen2020learning}, none
had been rigorously evaluated in practice until now. Thus, Our goal
is thus to assess their impact on performance.

In the following sections, we first review related work in game-playing
algorithms, with a focus on search methods and recent advances such
as AlphaZero and Athénan (Section~\ref{sec:Related-Work}). We then
detail the four modifications of the Unbounded Best-First Minimax
algorithm that are the focus of this study (Section~\ref{sec:Modifications-to-Unbounded}).
Next, we present our experimental methodology and results, analyzing
the impact of each modification individually and in combination (Section~\ref{sec:Experimental-Study}).
Finally, we conclude with a summary of our findings and discuss potential
directions for future research (Section~\ref{sec:Conclusion}).

\section{Related Work\label{sec:Related-Work}}

Game-playing algorithms \cite{bouzy2020artificial,millington2009artificial}
have long been a central focus of artificial intelligence, particularly
in the study of two-player zero-sum games. In this section, we first
review the main families of search algorithms used in such adversarial
settings, and then discuss a number of key techniques that have been
developed to improve their performance.

\subsection{Search Algorithms for Games}

The two most established approaches to adversarial search are Minimax
with Alpha-Beta pruning~\cite{knuth1975analysis} and Monte Carlo
Tree Search (MCTS)~\cite{Coulom06,browne2012survey}. 

Minimax with Alpha-Beta pruning is a depth-limited, exhaustive search
strategy that prunes subtrees which cannot influence the final decision.
Its time complexity is exponential in the search depth, making it
effective primarily when the branching factor is small or when deep
search is made feasible through domain-specific optimizations. Despite
this, Alpha-Beta remains the backbone of many traditional chess engines
due to its deterministic behavior and compatibility with various enhancements.
The most famous achievement based on Alpha-Beta is the Deep Blue Chess
program \cite{campbell2002deep}, first program to beat a world champion
at Chess. 

Monte Carlo Tree Search, on the other hand, has emerged as a powerful
alternative, particularly since its use in AlphaGo~\cite{silver2016mastering}
and later AlphaZero. Like the Unbounded Best-First Minimax algorithm,
MCTS performs an iterative search by traversing the game tree down
to a leaf at each iteration. However, the traversal policy in MCTS
is based not on the heuristic minimax value, but on a balance between
exploitation and exploration: it selects the action that maximizes
the average evaluation (win rate) of the subtree, biased by an exploration
term. This term typically follows an upper confidence bound (UCB)
formula, favoring less-explored actions to ensure a broader search.
In AlphaZero, the win percentage is estimated by a neural network,
and the exploration term is biased by the learned policy.

The Unbounded Best-First Minimax algorithm~\cite{cohen2025study,cohen2020learning,korf1996best}---central
to the Athénan system---is an alternative that lies between classical
Minimax and MCTS. It conducts a focused search by always expanding
the most promising sequence according to current minimax estimates.
Unlike Alpha-Beta and like MCTS, it is an anytime algorithm. It does
not set a maximum search depth. More precisely, unlike Alpha-Beta,
it does not explore uniformly to a given depth. In addition, unlike
MCTS, it does not rely on stochastic simulations, average evaluations
or exploration terms. As a result, UBFM can be highly efficient in
time-limited scenarios and particularly effective when combined with
evaluation functions generated by reinforcement learning, notably
with Athénan~\cite{cohen2025study} .

Other search algorithms exist~\cite{bouzy2020artificial,cohen2025study,millington2009artificial}.
Let us mention in particular the following algorithms.

On the one hand, Memory Test Driver~\cite{plaat1996best} is an algorithm
that repeatedly calls Alpha-Beta search with a very narrow \emph{search
window}, which allows for extremely aggressive pruning. After each
call, it uses the result to adjust the search bounds and repeat the
process until it identifies the exact value of the root state with
minimal computation.

On the other hand, Principal Variation Search~\cite{pearl1980scout}
is a variant of Alpha-Beta that optimizes the search. The algorithm
assumes the first child is best and uses faster, shallow searches
for the others; if any seem equally good or better, a full search
is performed to confirm.

There are also hybridizations of the Minimax algorithm and the Monte
Carlo Tree Search algorithm~\cite{baier2018mcts}.

\subsection{Search Enhancements and Optimizations}

Several important techniques have been developed to enhance the performance
of search algorithms, especially in fixed-depth search frameworks
like Alpha-Beta. 

One of the most widely used is \emph{move ordering}~\cite{fink1982enhancement},
which attempts to place promising moves earlier in the search order.
Good move ordering increases the likelihood of pruning and can dramatically
reduce the number of evaluated nodes.

Another foundational technique is \emph{iterative deepening}~ \cite{korf1985depth},
which transforms fixed-depth search into an anytime algorithm by repeatedly
running depth-limited searches with increasing depth. This not only
allows the algorithm to return a best-so-far move at any time but
also improves move ordering across iterations by using information
gathered from shallower searches.

The $k$-best pruning technique \cite{baier2018mcts} is used to reduce
the exponential growth of the search space with respect to the number
of anticipated rounds. Since the number of states grows exponentially
with the branching factor, the method limits the search to the best
actions per state. These actions are estimated using the results of
the previous search. While this significantly speeds up the search
and allows deeper lookahead, it introduces inaccuracy, as optimal
strategies might lie in the pruned branches. Moreover, the technique
relies on a tunable parameter, $k$, which must be adapted to each
domain. 

Transposition tables~\cite{greenblatt1988greenblatt} are another
common enhancement, used to avoid redundant evaluation of identical
game states reached via different sequences of actions. By turning
the tree into a directed acyclic graph (DAG), they significantly reduce
the effective size of the search space. In addition to their use in
the context of Alpha-Beta, transposition tables have also been used
with MCTS~\cite{browne2012survey}.

In the context of MCTS, work has proposed incorporating state resolution
into the search process~\cite{winands2008monte,cazenave2021monte}.
For instance, a state known to be a win or a loss (due to exhaustive
search) can be prioritized in selection and backpropagation, thus
improving decision quality. This mirrors ideas present in the completition
technique used in UBFM.

Overviews of MCTS improvements were conducted~\cite{swiechowski2023monte,browne2012survey}.

\section{Modifications to Unbounded Best-First Minimax\label{sec:Modifications-to-Unbounded}}

In this section, we describe the four algorithmic modifications evaluated
in this work. Each of them alters a distinct component of the original
Unbounded Best-First Minimax algorithm. 

\subsection{Use of Transposition Tables}

Transposition tables store information about previously encountered
game states, such as evaluations, in a dictionary to avoid redundant
computation. When applied this technique transforms the game search
structure from a tree into a directed acyclic graph, enabling the
reuse of subtrees that are reachable via different action sequences.
Formalization of Unbounded Best-First Minimax with transposition table
is available in Algorithm 1 of \cite{cohen2020learning}.

Recall that although transposition tables are a standard tool in Alpha-Beta
search, the impact of transpositions in the context of Unbounded Best-First
Minimax has never been evaluated until now.

\subsection{Backpropagation Strategy: Original vs. Full}

The original UBFM algorithm, as described by Korf and Chickering~\cite{korf1996best},
performs value backpropagation after each leaf expansion by climbing
up the game tree only until the value of a node remains unchanged.
Let's call this node the \emph{invariant node.} It then start the
new iteration from this node. This technique limits the amount of
work per iteration in relation to the formalization of Cohen-Solal~\cite{cohen2020learning}.
However, the two algorithms do not build the same search tree in the
presence of value ties or transposition tables.

In this second formalization, the first state of each next iteration
is different. All iteration restart from the root tree. Moreover,
backpropagation is performed up to the root, even if the value of
a state encountered does not change. Therefore, on the one hand, when
the first equality is encountered (if the corresponding node is higher
in the tree than the last invariant node), the tiebreaker can lead
to extending a different action sequence. On the other hand, by starting
from the root, we can encounter at the top of the tree a game state
that we encountered at the bottom of the tree during the previous
iteration. If the transposition tables are used, the value of this
state at the top of the tree has been updated, which can lead to a
different decision at this state by restarting from the root. So this
may lead to a new different iteration. As a result, the Korf \& Chickering
algorithm searches more deeply in these cases than the algorithm of
Cohen-Solal. In fact, more generally, by backpropagating values in
the game tree to the root, it can even bring up the values of states
higher in the tree than the invariant node. This occurs when the value
of a state has been updated using transposition tables, and this update
causes the value of its parent states to change. Such an update of
the parent states would therefore not have been performed with the
Korf \& Chickering algorithm.

It is not clear which formalization is better than the other or even
if there is a real difference in practice. a This difference has never
been evaluated in practice.

\subsection{Learned Evaluation Function vs. Terminal Evaluation}

The standard use of search algorithms uses a heuristic evaluation
function for non-terminal states and an exact evaluation function
for terminal states. Recall that terminal states are the endgame states.
Evaluations of such states are thus the true outcome of end-game states
(e.g., win, loss, draw or more generally the game score). Since the
heuristic evaluation function is an approximation of the exact evaluation
function, it is natural to use the exact evaluation function. However,
in some cases, the exact evaluation function is more expensive than
the heuristic evaluation function. It may be useful in such a case
to replace the exact evaluation function with the heuristic evaluation
function. If using the heuristic function does not change performance
or only slightly degrades it, then it may be worthwhile to use this
modification. Conversely, if it significantly degrades performance,
such a modification should not be considered. We will investigate
in this article whether such a change might be of interest.

\subsection{Completion}

Completion is a technique designed to improve decision quality by
explicitly distinguishing between resolved and unresolved states during
search. A state is considered resolved when its value is known with
certainty (thanks to the properties of min and max and a sufficiently
extensive search tree). Otherwise, it is said unresolved.

With the completion technique, UBFM prioritizes resolved states in
decision making. Specifically, it always selects a resolved winning
state over an unresolved one, even if the unresolved state has a higher
heuristic value. In addition, it avoids selecting resolved losing
states. Furthermore, it never chooses a solved state when searching.
Without completion, Unbounded Best-First Minimax does not always compute
the winning strategy when it exists (the basic algorithm can get stuck
in local fixpoints)~\cite{cohen2021completeness}.

Formalization of Unbounded Best-First Minimax with completion is available
in Algorithm 12 of \cite{cohen2020learning}.

\section{Experimental Study\label{sec:Experimental-Study}}

We conducted a empirical evaluation of the four modifications to the
Unbounded Best-First Minimax algorithm described in the previous section.
This study aims to quantify their individual and combined impact on
search performance, using standardized benchmarks across several two-player
games.

\subsection{Experimental Protocol}

We apply the same experimental protocol for evaluating search algorithms
as the article ~\cite{cohen2025study}. We use the same games, the
same evaluation functions, the same procedures, etc. All the details
of this evaluation are therefore available in \cite{cohen2025study}.
So we are content with a summary. Note that, the performance of the
algorithms evaluated in this previous study is therefore directly
comparable with the performance of the algorithms in this study.

We evaluated each algorithmic variant in the context of 22 deterministic,
perfect-information, zero-sum games, namelly Amazons, Ataxx, Breakthrough,
Brazilian Draughts, Canadian DraughtsClobber, Connect6, International
Draughts, Chess, Go $9\times9$, Outer-OpenGomoku, Havannah, Hex $11\times11$,
Hex $13\times13,$ Hex $19\times19$, Lines of Action, Othello $10\times10$,
Othello $8\times8$, Santorini, Surakarta and, Xiangqi. There are
all recurrent games at the Computer Olympiad, the global artificial
intelligence board game competition.  

Each algorithm is evaluated against the so-called \emph{benchmark
adversary Unbounded Minimax~\cite{cohen2025study}} corresponding
to Unbounded Minimax with transposition tables, completion, exact
terminal evaluation, and full backpropagation.

For each game, six distinct sets of evaluation functions are used,
each containing approximately fifteen functions. The evaluation process
is repeated separately for each of these six sets. During each repetition,
every search algorithm plays against the benchmark opponent twice
(once as the first player and once as the second) for every pair of
evaluation functions in the current set. This results in roughly 450
matches per algorithm per repetition, totaling about 2,700 matches
per game across all six repetitions. Each search algorithm uses 10
seconds of search time per action.

To assess the performance of a search algorithm, match outcomes are
aggregated using the following scoring: 1 for a win, -1 for a loss,
and 0 for a draw. The final performance on a game is the average score
across all matches played by the algorithm in that game. We report
5\% confidence intervals for these values, along with an overall average
performance across all games. In addition, we provide an unbiased
global confidence interval using stratified bootstrapping at the 5\%
level.

For clarity and compactness in data tables, all game-specific performance
percentages are rounded to the nearest whole number.

\subsection{Results}

We evaluate the following algorithms: benchmark Unbounded Minimax
$\ubfmref$, benchmark Unbounded Minimax without completion $\ubfmnoc$,
benchmark Unbounded Minimax without transposition table $\ubfmnott$,
benchmark Unbounded Minimax with Korf \& Chickering backpropagation
$\ubfmkcback$, benchmark Unbounded Minimax without without completion
nor exact terminal evaluation $\ubfmnocete$.

The results are summarized in Table~\ref{tab:Performance-variante-UBFM}. 

{\small{}
\begin{table}
{\small\centering{}{}\caption{Performance of the different studied variants of Unbounded Best-First
Minimax\label{tab:Performance-variante-UBFM}}
}{\small{}%
\begin{tabular}{ccccccc}
\cline{2-5}
 &  & {\tiny{}mean} & {\tiny{}lower bound} & {\tiny{}upper bound} & \multicolumn{2}{c}{}\tabularnewline
\cline{2-5}
 & $\ubfmref$ & {\tiny -0.09} & {\tiny -0.51} & {\tiny -0.69} & \multicolumn{2}{c}{}\tabularnewline
 & $\ubfmnoc$, & {\tiny -5.03} & {\tiny -5.63} & {\tiny -4.43} & \multicolumn{2}{c}{}\tabularnewline
 & $\ubfmnott$ & {\tiny -4.15} & {\tiny -4.74} & {\tiny -3.55} & \multicolumn{2}{c}{}\tabularnewline
 & $\ubfmkcback$ & {\tiny -2.37} & {\tiny -2.97} & {\tiny -1.78} & \multicolumn{2}{c}{}\tabularnewline
 & $\ubfmnocete$ & {\tiny -13.13} & {\tiny -13.71} & {\tiny -12.56} & \multicolumn{2}{c}{}\tabularnewline
\hline 
 & {\tiny{}Amazons} & {\tiny{}Arimaa} & {\tiny{}Ataxx} & {\tiny{}Breakthrough} & {\tiny{}Brazilian} & {\tiny{}Canadian}\tabularnewline
\hline 
$\ubfmref$ & {\tiny -1\% \textpm 3\%} & {\tiny 0\% \textpm 1\%} & {\tiny 5\% \textpm 5\%} & {\tiny -1\% \textpm 3\%} & {\tiny 1\% \textpm 1\%} & {\tiny 0\% \textpm 3\%}\tabularnewline
$\ubfmnoc$, & {\tiny 0\% \textpm 3\%} & {\tiny -11\% \textpm 2\%} & {\tiny -27\% \textpm 4\%} & {\tiny 0\% \textpm 3\%} & {\tiny -4\% \textpm 1\%} & {\tiny -12\% \textpm 3\%}\tabularnewline
$\ubfmnott$ & {\tiny -1\% \textpm 3\%} & {\tiny -8\% \textpm 1\%} & {\tiny -30\% \textpm 4\%} & {\tiny 0\% \textpm 3\%} & {\tiny 0\% \textpm 1\%} & {\tiny -7\% \textpm 3\%}\tabularnewline
$\ubfmkcback$ & {\tiny 0\% \textpm 3\%} & {\tiny -5\% \textpm 1\%} & {\tiny -33\% \textpm 4\%} & {\tiny -1\% \textpm 3\%} & {\tiny -1\% \textpm 1\%} & {\tiny -5\% \textpm 3\%}\tabularnewline
$\ubfmnocete$ & {\tiny -1\% \textpm 3\%} & {\tiny -33\% \textpm 2\%} & {\tiny -25\% \textpm 4\%} & {\tiny -3\% \textpm 3\%} & {\tiny -5\% \textpm 1\%} & {\tiny -16\% \textpm 3\%}\tabularnewline
\hline 
 & {\tiny{}Clobber} & {\tiny{}Connect6} & {\tiny{}International} & {\tiny{}Chess} & {\tiny{}Go 9} & {\tiny{}Gomoku}\tabularnewline
\hline 
$\ubfmref$ & {\tiny -2\% \textpm 3\%} & {\tiny 2\% \textpm 5\%} & {\tiny 0\% \textpm 1\%} & {\tiny 1\% \textpm 2\%} & {\tiny 0\% \textpm 3\%} & {\tiny 0\% \textpm 3\%}\tabularnewline
$\ubfmnoc$, & {\tiny 0\% \textpm 3\%} & {\tiny -5\% \textpm 5\%} & {\tiny -5\% \textpm 1\%} & {\tiny 0\% \textpm 2\%} & {\tiny -1\% \textpm 3\%} & {\tiny 2\% \textpm 2\%}\tabularnewline
$\ubfmnott$ & {\tiny -3\% \textpm 3\%} & {\tiny -14\% \textpm 5\%} & {\tiny -2\% \textpm 1\%} & {\tiny -1\% \textpm 2\%} & {\tiny -4\% \textpm 3\%} & {\tiny -4\% \textpm 2\%}\tabularnewline
$\ubfmkcback$ & {\tiny 0\% \textpm 3\%} & {\tiny 1\% \textpm 5\%} & {\tiny -2\% \textpm 1\%} & {\tiny -4\% \textpm 2\%} & {\tiny -1\% \textpm 3\%} & {\tiny 1\% \textpm 3\%}\tabularnewline
$\ubfmnocete$ & {\tiny 1\% \textpm 3\%} & {\tiny -2\% \textpm 5\%} & {\tiny -7\% \textpm 1\%} & {\tiny -19\% \textpm 2\%} & {\tiny -37\% \textpm 3\%} & {\tiny 0\% \textpm 2\%}\tabularnewline
\cline{1-6}
 & {\tiny{}Havannah} & {\tiny{}Hex 11} & {\tiny{}Hex 13} & {\tiny{}Hex 19} & {\tiny{}Lines of A.} & \tabularnewline
\cline{1-6}
$\ubfmref$ & {\tiny -2\% \textpm 3\%} & {\tiny 0\% \textpm 3\%} & {\tiny 2\% \textpm 3\%} & {\tiny -2\% \textpm 3\%} & {\tiny 2\% \textpm 2\%} & \tabularnewline
$\ubfmnoc$, & {\tiny 0\% \textpm 3\%} & {\tiny 1\% \textpm 3\%} & {\tiny 0\% \textpm 3\%} & {\tiny -1\% \textpm 3\%} & {\tiny -1\% \textpm 2\%} & \tabularnewline
$\ubfmnott$ & {\tiny -3\% \textpm 3\%} & {\tiny -2\% \textpm 3\%} & {\tiny -3\% \textpm 3\%} & {\tiny -2\% \textpm 3\%} & {\tiny -1\% \textpm 2\%} & \tabularnewline
$\ubfmkcback$ & {\tiny 0\% \textpm 3\%} & {\tiny 1\% \textpm 3\%} & {\tiny -1\% \textpm 3\%} & {\tiny 0\% \textpm 3\%} & {\tiny 1\% \textpm 2\%} & \tabularnewline
$\ubfmnocete$ & {\tiny 1\% \textpm 3\%} & {\tiny 1\% \textpm 3\%} & {\tiny 1\% \textpm 3\%} & {\tiny 2\% \textpm 3\%} & {\tiny -7\% \textpm 2\%} & \tabularnewline
\cline{1-6}
 & {\tiny{}Othello 10} & {\tiny{}Othello 8} & {\tiny{}Santorini} & {\tiny{}Surakarta} & {\tiny{}Xiangqi} & \tabularnewline
\cline{1-6}
$\ubfmref$ & {\tiny 1\% \textpm 3\%} & {\tiny -1\% \textpm 2\%} & {\tiny 0\% \textpm 5\%} & {\tiny 0\% \textpm 3\%} & {\tiny 1\% \textpm 3\%} & \tabularnewline
$\ubfmnoc$, & {\tiny -15\% \textpm 3\%} & {\tiny -17\% \textpm 3\%} & {\tiny -7\% \textpm 5\%} & {\tiny -7\% \textpm 3\%} & {\tiny -11\% \textpm 3\%} & \tabularnewline
$\ubfmnott$ & {\tiny -7\% \textpm 3\%} & {\tiny 0\% \textpm 2\%} & {\tiny -7\% \textpm 5\%} & {\tiny -1\% \textpm 3\%} & {\tiny -6\% \textpm 3\%} & \tabularnewline
$\ubfmkcback$ & {\tiny -2\% \textpm 3\%} & {\tiny -3\% \textpm 2\%} & {\tiny -7\% \textpm 5\%} & {\tiny -3\% \textpm 3\%} & {\tiny -2\% \textpm 3\%} & \tabularnewline
$\ubfmnocete$ & {\tiny -13\% \textpm 3\%} & {\tiny -14\% \textpm 3\%} & {\tiny -10\% \textpm 5\%} & {\tiny -17\% \textpm 3\%} & {\tiny -89\% \textpm 1\%} & \tabularnewline
\end{tabular}}
\end{table}
}{\small\par}

These results indicate that each modification has a measurable impact
on the algorithm\textquoteright s behavior. Completion, transposition
tables, and full backpropagation increase the performance. These improvements
must therefore be considered when putting Unbounded Best-First Minimax
into practice.

Not using exact terminal evaluation slightly decreases performance.
Therefore, we recommend using exact terminal evaluation, except when
the terminal evaluation function is particulary expensive. In this
case, not using it should have a very significant positive impact.

\section{Conclusion\label{sec:Conclusion}}

In this paper, we conducted the first systematic empirical study of
four modifications to the Unbounded Best-First Minimax algorithm.
This search algorithm, recently brought to prominence by the Athénan
algorithm, offers an efficient alternative to classical game tree
search methods such as Alpha-Beta pruning and Monte Carlo Tree Search.

Our experimental results demonstrate that each evaluated modification
has a distinct and measurable impact on performance. In particular,
the use of transposition tables, full backpropagation, and completion
provides significant improvements. 

Furthermore, our study suggests that not using an exact terminal evaluation
function may be useful, but only when its computation is expensive.

\bibliographystyle{plain}
\bibliography{jeux}

\begin{thebibliography}{10}

\bibitem{baier2018mcts}
Hendrik Baier and Mark~HM Winands.
\newblock Mcts-minimax hybrids with state evaluations.
\newblock {\em Journal of Artificial Intelligence Research}, 62:193--231, 2018.

\bibitem{bouzy2020artificial}
Bruno Bouzy, Tristan Cazenave, Vincent Corruble, and Olivier Teytaud.
\newblock Artificial intelligence for games.
\newblock {\em A Guided Tour of Artificial Intelligence Research: Volume II: AI
  Algorithms}, pages 313--337, 2020.

\bibitem{browne2012survey}
Cameron~B Browne, Edward Powley, Daniel Whitehouse, Simon~M Lucas, Peter~I
  Cowling, Philipp Rohlfshagen, Stephen Tavener, Diego Perez, Spyridon
  Samothrakis, and Simon Colton.
\newblock A survey of monte carlo tree search methods.
\newblock {\em Transactions on Computational Intelligence and AI in games},
  4(1):1--43, 2012.

\bibitem{campbell2002deep}
Murray Campbell, A~Joseph Hoane~Jr, and Feng-hsiung Hsu.
\newblock Deep blue.
\newblock {\em Artificial Intelligence}, 134(1-2):57--83, 2002.

\bibitem{cazenave2021monte}
Tristan Cazenave.
\newblock Monte carlo game solver.
\newblock In {\em Monte Carlo Search: First Workshop, MCS 2020, Held in
  Conjunction with IJCAI 2020, Virtual Event, January 7, 2021, Proceedings 1},
  pages 56--70. Springer, 2021.

\bibitem{cohen2020learning}
Quentin Cohen-Solal.
\newblock Learning to play two-player perfect-information games without
  knowledge.
\newblock {\em arXiv preprint arXiv:2008.01188}, 2020.

\bibitem{cohen2021completeness}
Quentin Cohen-Solal.
\newblock Completeness of unbounded best-first game algorithms.
\newblock {\em arXiv preprint arXiv:2109.09468}, 2021.

\bibitem{cohen2025study}
Quentin Cohen-Solal.
\newblock Study and improvement of search algorithms in two-players perfect
  information games.
\newblock {\em arXiv preprint}, 2025.

\bibitem{cohen2023minimax}
Quentin Cohen-Solal and Tristan Cazenave.
\newblock Minimax strikes back.
\newblock {\em AAMAS}, 2023.

\bibitem{Coulom06}
R{\'{e}}mi Coulom.
\newblock Efficient selectivity and backup operators in monte-carlo tree
  search.
\newblock In {\em Computers and Games, 5th International Conference, {CG} 2006,
  Turin, Italy, May 29-31, 2006. Revised Papers}, pages 72--83, 2007.

\bibitem{fawzi2022discovering}
Alhussein Fawzi, Matej Balog, Aja Huang, Thomas Hubert, Bernardino
  Romera-Paredes, Mohammadamin Barekatain, Alexander Novikov, Francisco~J
  R~Ruiz, Julian Schrittwieser, Grzegorz Swirszcz, et~al.
\newblock Discovering faster matrix multiplication algorithms with
  reinforcement learning.
\newblock {\em Nature}, 610(7930):47--53, 2022.

\bibitem{fink1982enhancement}
William Fink.
\newblock An enhancement to the iterative, alpha-beta, minimax search
  procedure.
\newblock {\em ICGA Journal}, 5(1):34--35, 1982.

\bibitem{greenblatt1988greenblatt}
Richard~D Greenblatt, Donald~E Eastlake, and Stephen~D Crocker.
\newblock The greenblatt chess program.
\newblock In {\em Computer chess compendium}, pages 56--66. Springer, 1988.

\bibitem{knuth1975analysis}
Donald~E Knuth and Ronald~W Moore.
\newblock An analysis of alpha-beta pruning.
\newblock {\em Artificial Intelligence}, 6(4):293--326, 1975.

\bibitem{korf1985depth}
Richard~E Korf.
\newblock Depth-first iterative-deepening: An optimal admissible tree search.
\newblock {\em Artificial Intelligence}, 27(1):97--109, 1985.

\bibitem{korf1996best}
Richard~E Korf and David~Maxwell Chickering.
\newblock Best-first minimax search.
\newblock {\em Artificial intelligence}, 84(1-2):299--337, 1996.

\bibitem{mankowitz2023faster}
Daniel~J Mankowitz, Andrea Michi, Anton Zhernov, Marco Gelmi, Marco Selvi,
  Cosmin Paduraru, Edouard Leurent, Shariq Iqbal, Jean-Baptiste Lespiau, Alex
  Ahern, et~al.
\newblock Faster sorting algorithms discovered using deep reinforcement
  learning.
\newblock {\em Nature}, 618(7964):257--263, 2023.

\bibitem{millington2009artificial}
Ian Millington and John Funge.
\newblock {\em Artificial intelligence for games}.
\newblock CRC Press, 2009.

\bibitem{pearl1980scout}
Judea Pearl.
\newblock Scout: A simple game-searching algorithm with proven optimal
  properties.
\newblock In {\em AAAI}, pages 143--145, 1980.

\bibitem{plaat1996best}
Aske Plaat, Jonathan Schaeffer, Wim Pijls, and Arie De~Bruin.
\newblock Best-first fixed-depth minimax algorithms.
\newblock {\em Artificial Intelligence}, 87(1-2):255--293, 1996.

\bibitem{silver2016mastering}
David Silver, Aja Huang, Chris~J Maddison, Arthur Guez, Laurent Sifre, George
  Van Den~Driessche, Julian Schrittwieser, Ioannis Antonoglou, Veda
  Panneershelvam, Marc Lanctot, et~al.
\newblock Mastering the game of go with deep neural networks and tree search.
\newblock {\em Nature}, 529(7587):484, 2016.

\bibitem{silver2018general}
David Silver, Thomas Hubert, Julian Schrittwieser, Ioannis Antonoglou, Matthew
  Lai, Arthur Guez, Marc Lanctot, Laurent Sifre, Dharshan Kumaran, Thore
  Graepel, et~al.
\newblock A general reinforcement learning algorithm that masters chess, shogi,
  and go through self-play.
\newblock {\em Science}, 362(6419):1140--1144, 2018.

\bibitem{silver2017mastering}
David Silver, Julian Schrittwieser, Karen Simonyan, Ioannis Antonoglou, Aja
  Huang, Arthur Guez, Thomas Hubert, Lucas Baker, Matthew Lai, Adrian Bolton,
  et~al.
\newblock Mastering the game of go without human knowledge.
\newblock {\em Nature}, 550(7676):354, 2017.

\bibitem{swiechowski2023monte}
Maciej {\'S}wiechowski, Konrad Godlewski, Bartosz Sawicki, and Jacek
  Ma{\'n}dziuk.
\newblock Monte carlo tree search: A review of recent modifications and
  applications.
\newblock {\em Artificial Intelligence Review}, 56(3):2497--2562, 2023.

\bibitem{winands2008monte}
Mark~HM Winands, Yngvi Bj{\"o}rnsson, and Jahn-Takeshi Saito.
\newblock Monte-carlo tree search solver.
\newblock In {\em International Conference on Computers and Games}, pages
  25--36. Springer, 2008.

\end{thebibliography}

\end{document}